\documentclass[sigconf]{acmart}

\AtBeginDocument{%
  \providecommand\BibTeX{{%
    \normalfont B\kern-0.5em{\scshape i\kern-0.25em b}\kern-0.8em\TeX}}}

\setcopyright{acmcopyright}
\copyrightyear{2022}
\acmYear{2022}
\acmDOI{10.1145/1122445.1122456}

\copyrightyear{2022}
\acmYear{2022}
\setcopyright{acmcopyright}\acmConference[SIGMOD '22]{Proceedings of the 2022 International Conference on Management of Data}{June 12--17, 2022}{Philadelphia, PA, USA}
\acmBooktitle{Proceedings of the 2022 International Conference on Management of Data (SIGMOD '22), June 12--17, 2022, Philadelphia, PA, USA}
\acmPrice{15.00}
\acmDOI{10.1145/3514221.3526129}
\acmISBN{978-1-4503-9249-5/22/06}

\usepackage[utf8]{inputenc}
\usepackage[T1]{fontenc}
\usepackage{amsfonts}
\usepackage{microtype}
\usepackage{hyperref}
\usepackage{tikz}
\usetikzlibrary{arrows,decorations.pathmorphing,backgrounds,positioning,fit,petri}
\usepackage{booktabs}
\usepackage{graphicx}
\usepackage{makecell}
\usepackage[ruled,vlined]{algorithm2e}
\usepackage{subcaption}
\usepackage{bm}
\usepackage{tabularx}

\usepackage{textcomp}
\usepackage{amsthm}
\usepackage{amsmath}
\usepackage{verbatim}

\newcommand{\R}{\mathbb{R}}

\newcommand{\revised}[1]{#1}

\begin{document}

\fancyhead{}

\title{CoLES: Contrastive Learning for Event Sequences with Self-Supervision}

\author{Dmitrii Babaev} \thanks{E-mail: dmitri.babaev@gmail.com}
\affiliation{
    \institution{AIRI}
    \institution{Sber AI Lab }
    \country{Moscow, Russia}}

\author{Nikita Ovsov} 
\author{Ivan Kireev} 
\author{Maria Ivanova}
\affiliation{
    \institution{Sber AI Lab}  
    \country{Moscow, Russia}}

    
\author{Gleb Gusev}
\affiliation{  
    \institution{Sber AI Lab} 
    \institution{MIPT} 
    \country{Moscow, Russia}} 

\author{Ivan Nazarov}
\affiliation{  
    \institution{AIRI}
    \country{Moscow, Russia}} 


\author{Alexander Tuzhilin}
\affiliation{
    \institution{New York University}
    \country{New York, USA}}

\begin{abstract}
We address the problem of self-supervised learning on discrete event sequences generated by real-world users. Self-supervised learning incorporates complex information from the raw data in low-dimensional fixed-length vector representations that could be easily applied in various downstream machine learning tasks. In this paper, we propose a new method ``CoLES'', which adapts contrastive learning, previously used for audio and computer vision domains, to the discrete event sequences domain in a self-supervised setting.

We deployed CoLES embeddings based on sequences of transactions at the large European financial services company. Usage of CoLES embeddings significantly improves the performance of the pre-existing models on downstream tasks and produces significant financial gains, measured in hundreds of millions of dollars yearly.
We also evaluated CoLES on several public event sequences datasets and showed that CoLES representations consistently outperform other methods on different downstream tasks.
\end{abstract}

\ccsdesc[300]{Information systems~Data management systems}
\ccsdesc[300]{Applied computing~Online banking}
\ccsdesc[500]{Computing methodologies~Machine learning algorithms}

\keywords{representation learning, metric learning, contrastive learning, self-supervised learning, event sequences, data management}

\maketitle

\section{Introduction} \label{sec-intro}

As part of representation learning methods, data embedding aims to represent the relevant
intrinsic patterns of the point or sequential data into low-dimensional fixed-length vectors
capturing its ``essence'', that are useful in related downstream tasks,%
\citep{Mikolov2013EfficientEO,Peters2018DeepCW,Devlin2019BERTPO,Dosovitskiy2014DiscriminativeUF,Oord2018RepresentationLW}.
As such, pre-trained embeddings in different domains are used either as informative
out-of-the-box input features for Machine Learning or Deep Learning models without extensive
engineering or deep domain knowledge on the part of practitioners, or as building blocks
in representations of composite multi-modal data. In big-data applications embeddings may
be viewed as a learnable task-aware data compression technique, which enables storage-efficient
data sharing arrangements, possibly with privacy guarantees depending on used method.

Most research and application of embedding methods, however, have been focused on the core
machine learning domains, including ELMO~\citep{Peters2018DeepCW} and BERT~\citep{Devlin2019BERTPO}
in the natural language processing (NLP), CPC~\citep{Oord2018RepresentationLW} in speech recognition,
and various methods in computer vision (CV)~\citep{Dosovitskiy2014DiscriminativeUF, Oord2018RepresentationLW}.

The common feature of these domains is that that the data in such modalities is \emph{context
sensitive}: a term can be accurately reconstructed from the context-conditional language
model, similarly to the way a pixel can be inferred from its neighborhood. This property
underlies popular approaches for representation learning in NLP, such as BERT's Cloze
task~\citep{Devlin2019BERTPO}, and in audio and CV, such as CPC~\citep{Oord2018RepresentationLW}.

However, not every sequential discrete data features high mutual information between a single
item and its immediate neighborhood. For example, log entries, IoT telemetry, industrial
maintenance, user behavior \citep{Ni2018PerceiveYU}, travel patterns, transactional data,
and other industrial and financial event sequences typically consist of interleaved relatively
independent sub-streams.
For example, the transactions generated either by individual or business customers feature
irregular and periodic patterns, seen from the perspective of the financial services company
as a stream of unlabelled and apparently unrelated events. The most state-of-the-art
representation learning methods for token and sequence embedding from the NLP or CV are
not guaranteed to capture the peculiarities of such financial data, which exhibits customer
behavior of a certain type and constitutes valuable information for the fraud prevention
and development of efficient financial products.

In this paper, we propose a novel self-supervised method for embedding discrete event
sequences, called \emph{COntrastive Learning for Event Sequences (CoLES)}, which is based
on contrastive learning, \citep{Xing2002DistanceML, Hadsell2006DimensionalityRB}, with
a special data augmentation strategy.
Contrastive learning aims to learn a representation $x \mapsto M(x)$, which brings
\emph{positive pairs}, i.e. semantically similar objects, \emph{closer} to each other
in the embedding space, while \emph{negative pairs}, i.e. dissimilar objects, \emph{further}
apart.
Positive-negative pairs are obtained either \emph{explicitly} from the known ground-truth
target data or \emph{implicitly} using \emph{self-supervised} data augmentation strategies
\citep{Falcon2020AFF}. In the latter, the most common approach is conditional generation:
for a given pair of distinct datapoints $x \neq y$ the positive pairs $(z, z')$ are sampled
from the product $
  p_+(z, z') = p(z \mid x) \,p(z' \mid x) 
$, while the negative pairs -- from $
  p_-(z, z') = p(z \mid x) \,p(z' \mid y)
$ for $x \neq y$, where $
  p(\cdot \mid x)
$ is a process sampling random augmentations of $x$.

As a self-supervised sequence embedding method, \emph{CoLES} uses a novel augmentation
algorithm, which generates sub-sequences of observed event sequences and uses them as
different high-dimensional views of the object (sequence) for contrastive learning.
The proposed generative process is specifically designed to address the observed interleaved
periodicity in financial transaction event sequences, which is the primary application of
our method (see Section~\ref{sec-commercial}).
Representations learnt by CoLES can be used as feature vectors
in supervised domain-related tasks \citep{Mikolov2013EfficientEO,Song2017LearningUE,Zhai2019LearningAU},
e.g. fraud detection or scoring tasks based on transaction history, or they can be fine-tuned
for out-of-domain tasks \citep{Yosinski2014HowTA}.

We have applied CoLES to four publicly available datasets of event sequences from different
domains, such as financial transactions, retail receipts and game assessment records. As we
show in the experimental section, CoLES produces representations, which achieve strong
performance results, comparable to the hand-crafted features produced by domain experts.
We also demonstrate that the fine-tuned CoLES representations consistently outperform
representations produces by alternative methods.
%
%
Additionally, we deployed CoLES embeddings in several applications in our organization and tested
the method against the models currently used in the company. Experimental results, demonstrate
CoLES embeddings significantly improve the performance of the pre-existing models on the downstream
tasks, which resulted in significant financial gains for the company.

This paper makes the following contributions. We
\begin{enumerate}
    \item present CoLES, a self-supervised method with a novel augmentation method, which adapts
    contrastive learning to the discrete event sequence domain;

    \item demonstrate that CoLES consistently outperforms existing supervised, self-supervised and
    semi-supervised learning baselines adapted to the event sequence domain;

    \item present the results of applying CoLES embeddings in the real-world scenaria and show
    that the proposed method can be of significant value for the day-to-day modelling in financial
    services industry.
\end{enumerate}

The rest of the paper is organized as follows. In the next section, we discuss related studies
on self-supervised and contrastive learning. In Section~\ref{sec-method}, we introduce our new
method CoLES for discrete event sequences. In Section~\ref{sec-exp}, we demonstrate that CoLES
outperforms several strong baselines including previously proposed contrastive learning methods
adapted to event sequence datasets. Section~\ref{sec-conclusions} is dedicated to the discussion
of our results and conclusions.
We provide the source code for all the experiments on public datasets described in this paper.\footnote{https://github.com/dllllb/coles-paper}

\begin{figure*}[htbp]
  \includegraphics[width=0.98\linewidth]{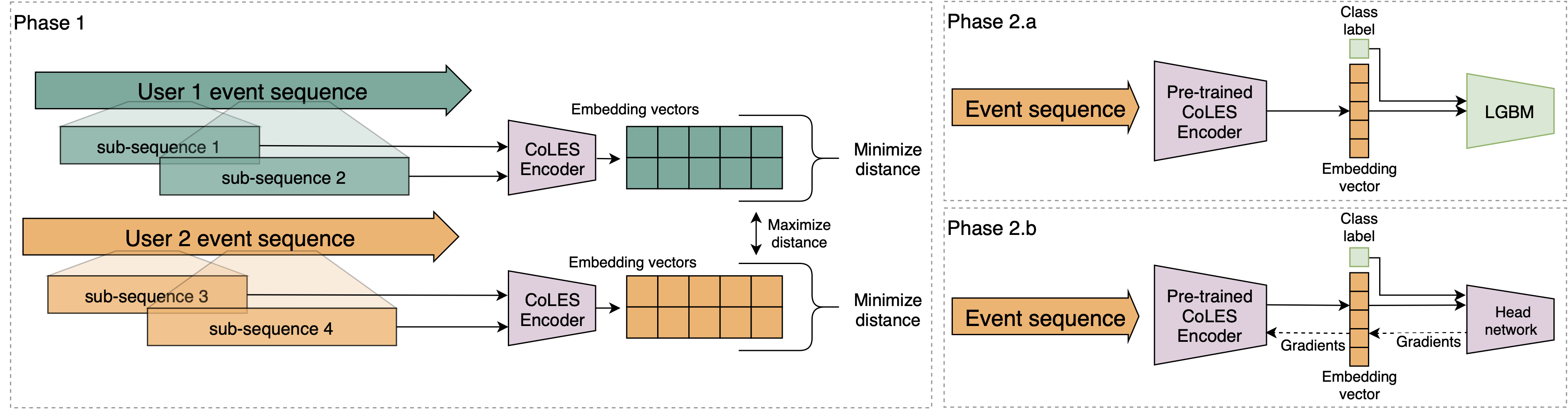}
    \caption{
        General framework.
        Phase 1: Self-supervised training.
        Phase 2.a Self-supervised embeddings as features for supevised model.
        Phase 2.b: Pre-trained encoder fine-tuning.
    }
  \label{fig-arch}
\end{figure*}

\section{Related work} \label{sec-rel-work}

Contrastive learning has been successfully applied to constructing low-dimensional representations
(embeddings) of various objects, such as images~\citep{Chopra2005LearningAS,Schroff2015FaceNetAU},
texts~\citep{Reimers2019SentenceBERTSE}, and audio recordings~\citep{Wan2018GeneralizedEL}.
Although the aim of these studies is to identify the object based on its sample
\citep{Schroff2015FaceNetAU,Hu2014DiscriminativeDM,Wan2018GeneralizedEL}, these supervised
approaches are not applicable to our setting, since their training datasets explicitly contain
multiple independent samples per each particular object, which form positive pairs as
a critical component for learning. 

For situations when positive pairs are not available or their amount is limited, synthetic data
generation and augmentation techniques could be employed. One of the first such frameworks was
proposed by \citet{Dosovitskiy2014DiscriminativeUF}, who introduce surrogate classes by randomly
augmenting the same image. Several recent works, e.g. \citep{Bachman2019LearningRB,He2019MomentumCF,Chen2020ASF},
extended this idea by applying contrastive learning methods (see \citep{Falcon2020AFF}).
%
Contrastive Predictive Coding (CPC) is a self-supervised learning approach proposed for non-discrete
sequential data in \citep{Oord2018RepresentationLW}. The CPC employs an autoregressive predictive
model of the input sequence in order to extract meaningful latent representations, and as such
can be adapted the domain of discrete event sequences (see Section~\ref{sec-res} for comparison
with CoLES).



Several publications consider self-supervision for user behavior sequences in the recommender system
and user behavior analysis domains. CPC-like approach for self-supervised learning on user click
histories is proposed in \citep{Zhou2020ContrastiveLF}, while \citep{Ma2020DisentangledSI} use
an auxiliary self-supervised sequence-to-sequence loss term.
In~\citep{Zhou2020S3RecSL}, it was proposed to use ``Cloze'' task from BERT~\citep{Devlin2019BERTPO}
for self-supervision on purchase sequences.
A SimCLR-like approach for text-based tasks and tabular data was adapted in \citep{Yao2020SelfsupervisedLF}.
\revised{
    In~\citep{Zhuang2019AttributedSE} authors propose an unsupervised autoregressive method
    to produce embeddings of attributed sequences where a sequence of categorical tokens has
    additional global attributes.
}
The aforementioned works consider sequences of ``items'', where each element is an item identifier.
We consider more complex sequences of events where an element of the sequence consists of several
categorical and numerical fields.

There are papers dedicated to supervised learning for discrete event sequences, e.g. \citep{Wiese2009CreditCT,Tobback2019RetailCS,Babaev2019ETRNNAD,chatterjee2003modeling,sinha2014your},
but self-supervised pre-training is not used in those works.


\section{Problem formulation and overview of the CoLES method} \label{sec-method}

\subsection{Problem formulation} \label{sec:problem setting}

While the method proposed in this study could be studied in different domains, we focus on discrete
sequences of events. Assume there are some entities $e$ and that each entity's lifetime activity is
observed as a sequence of events $
    x_e := \{x_e(t)\}^{T_e}_{t=1}
$. Entities could be people or organizations or some other abstractions. Events $x_e(t)$ may have
any nature and structure (e.g., transactions of a client, click logs of a user), and their components
may contain numerical, categorical, and textual fields (see datasets description in Section~\ref{sec-exp}). 

According to theoretical framework of contrastive learning proposed in \citep{Saunshi2019ICML}, each
entity $e$ is a latent class, which is associated with a distribution $P_e$ over its possible samples
(event sequences). However, unlike the problem setting of \cite{Saunshi2019ICML}, we have no positive
pairs, i.e. pairs of event sequences representing the same entity $e$. Instead, only one sequence
$x_e$ is available for the entity~$e$. Formally, each entity $e$ is associated with a latent stochastic
process $
    \{X_e(t)\} = \{X_e(t)\}_{t\geq 1}
$, and we observe \emph{only a single} finite realisation $
    \{x_e\} = \{x_e(t)\}_{t=1}^{T_e}
$ of it. Our goal is to learn an \emph{encoder} $M$ that maps event sequences into a feature space~$\R^d$
in such a way that the obtained \emph{embedding} $
    \{x_e\} \mapsto c_e = M(\{x_e\}) \in \R^d
$ encodes the essential properties of $e$ and disregards irrelevant noise contained in the sequence.
That is, the embeddings $M(\{x'\})$ and $M(\{x''\})$ should be close to each other, if $x'$ and
$x''$ were paths generated by the same process $\{X_e(t)\}$, and further apart, if generated by distinct
processes.

The quality of representations can be examined by downstream tasks in the two ways:
\begin{enumerate}
    \item $c_e$ can be used as a feature vector for a task--specific model (see Figure~\ref{fig-arch}, Phase 2a),
    \item encoder $M$ can also be (jointly) fine-tuned~\citep{Yosinski2014HowTA} (see Figure~\ref{fig-arch}, Phase 2b).
\end{enumerate}

\subsection{Sampling of surrogate sequences as an augmentation procedure} \label{sec-pos-pairs}

When there is no sampling access to the latent processes $\{X_e(t)\}$, one could employ synthetic
augmentation strategies, which are akin to bootstrapping.
\revised{
Most augmentation techniques proposed
for continuous domains, such as image displacement, color jitter or random gray scale in CV, \citep{Falcon2020AFF},
are not applicable to discrete events. Thus, generating \emph{sub-sequences} from the same event
sequence $\{x_e(t)\}$, could be used as a possible augmentation.
The idea proposed below resembles the bootstrap method, \citep{Efron1994Bootstrap}, which, roughly,
posits that \emph{the empirical distribution} induced by the observed sample of \emph{independent}
datapoints is a suitable proxy for the \emph{distribution of the population}.
In our setting, however, the events are not independent observations, which prompts us to
rely on different data assumptions.
}

The key property of event sequences that represent lifetime activity is periodicity and repeatability
of its events (see Section~\ref{sec-period} for the empirical observations of these properties
for the considered datasets). This motivates the \emph{Random slices} sampling method applied
in CoLES, as presented in Algorithm~\ref{alg-slce-ss}.
\revised{
    Sub-sequences $\{\tilde{x}_e\}$ are sampled
    from a given sequence $\{x_e(t)\}$ as continuous segments, ``slices'', using the following three steps.
    First, the length of the slice is chosen uniformly from the admissible values.
    Second, too short (and, optionally, too long) sub-sequences are discarded.
    Third, the starting position is uniformly chosen from all possible values.
}
The overview of the CoLES method is presented in Figure \ref{fig-arch}.

\begin{algorithm}
    \SetAlgoLined
    \textbf{hyperparameters:}
        $m, M$: minimal and maximal possible length of a sub-sequence;
        $k$: number of samples.
    \\ 
    \textbf{input:}
        A sequence $S = \{z_j\}_{j=0}^{T-1}$ of length $T$.
    \\
    \textbf{output:}
        $\mathcal{S}$: sub-sequences of $S$.
    \\
    \BlankLine
    \For{$i\leftarrow 1$ \KwTo $k$}{
        Generate a random integer $T_i$ uniformly from $[1, T]$;
        \\
        \uIf{$T_i\in [m, M]$}{  
            Generate a random integer $s$ from $[0, T - T_i)$;
            \\
            Add the slice $
                \tilde{S}_i := \bigl\{z_{s + j}\bigr\}_{j=0}^{T_i - 1}  
            $ to $\mathcal{S}$;
            \\
        }
    }
    \caption{Random slices sub-sequence generation strategy}
    \label{alg-slce-ss}
\end{algorithm}

\subsection{Model training} \label{sec-training}

\textbf{Batch generation.} The following procedure creates a batch for CoLES. $N$ initial
sequences are randomly taken and $K$ sub-sequences are produced for each of one. Pairs of
sub-sequences of from same sequence are used as positive samples and pairs from different
sequences -- as negatives.

\revised{
In the experiment section we consider several baseline empirical strategies for the sub-sequence
generation to compare with Algorithm~\ref{alg-slce-ss}. The details of the comparison are presented
in Section~\ref{sec-ablation}.
}


\textbf{Contrastive loss}
\revised{
We consider a classical variant of the contrastive loss, proposed in
\citep{Hadsell2006DimensionalityRB}, which minimizes the objective
\begin{equation*}
    \mathcal{L}_{u v}(M)
        = Y_{u v} \frac12 d_M(u, v)^2
        + (1 - Y_{u v}) \frac12\max\{0, \rho - d_M(u, v) \}^2
    \,,
\end{equation*}
with respect to $
    M \colon \mathcal{X} \to \mathbb{R}^n
$, where $
    d_M(u, v) = d(c_u, c_v)
$ is the distance between embeddings of the pair $(u, v)$, $
    c_* = M(\{\tilde{x}_*(\tau)\})
$, $Y_{u v}$ is a binary variable
identifying whether the pair $(u, v)$ is positive, and $\rho$ is the soft minimal margin between
dissimilar objects. The second term encourages separation of the embeddings in negative
pairs and prevents \emph{mode collapse} in $M$, when the entities are mapped to the same
point in the embedding space. $d(a, b)$ is the Euclidean distance, $
    d(a, b)
        = \sqrt{
            \sum_k (a_k - b_k)^2
        }
$, as proposed in~\citep{Hadsell2006DimensionalityRB}.
The sequences $\{\tilde{x}_u(\tau)\}$ and $\{\tilde{x}_v(\tau)\}$ of a pair with $Y_{u v} = 1$
are obtained through random slice generation (Algorithm~\ref{alg-slce-ss}) from the \emph{same}
observation $\{x_e(t)\}$, while in pairs with $Y_{u v} = 0$ the sequences are sampled from
$\{x_e(t)\}$ and $\{x_g(t)\}$, respectively, for $e \neq g$.

In the experiment section we compare the basic variant of the contrastive loss
with alternative variants. The results of the comparison is presented in Section~\ref{sec-ablation}.
}

\textbf{Negative sampling.} One challenge in contrastive learning approach is that positive pairs
are overwhelmed by potential negative pairs. Furthermore, some of negative pairs are distant enough,
to not provide any valuable feedback through $\mathcal{L}$ to $M$ during training, \citep{SimoSerra2015DiscriminativeLO,Schroff2015FaceNetAU}. We compare the common negative sampling
methods in Section~\ref{sec-res}. Due to certain negative sampling approaches begin distance-aware
and in order to make the overall distance computation less inefficient we restrict the encoder $M$
to the class of maps that output unit-norm vectors in $\R^d$. Therefore the pairwise distance $
    d_M(u, v)^2
$ is just $2 - 2 c_u^\top c_v$, which only requires pairwise dot products between embeddings $
    c_v = M(\{\tilde{x}_u(t)\})
$.

\subsection{Encoder architecture} \label{sec-enc-arch}

Embedding a sequence of events into a vector of fixed size requires encoding individual events
followed by aggregating the entire sequence. The composite encoder model $M$ in CoLES is of
the form $
    M(\{x_t\})
        := \phi_{\mathrm{seq}}(
            \{\phi_{\mathrm{evt}}(x_t)\}
        )
$, where $\phi_{\mathrm{evt}}$ and $\phi_{\mathrm{seq}}$ are event- and sequence-level
embedding networks, respectively, trained in an end-to-end manner to minimize the contrastive
loss  $\mathcal{L}(M)$.

\textbf{The event encoder} $\phi_{\mathrm{evt}}$ takes a set of attributes of each event $x_t$
and outputs its intermediate representation in $\R^d$: $z_t = \phi_{\mathrm{evt}}(x_t)$. This
encoder consists of several linear layers, for embedding one-hot encoded categorical attributes,
and batch normalization layers, applied to numerical attributes of events. Outputs of these
layers are concatenated to produce the event embedding $z_t$.

\textbf{The sequence encoder} $\phi_{\mathrm{seq}}$ takes the intermediate representations of
the events $ z_{1:T} = z_1, z_2, \cdots z_T $ and outputs the representation $c_t$ of their
sequence up to the time $t$: $c_t = \phi_{\mathrm{seq}}(z_{1:t})$. The last $c_T$ is used
as the embedding of the whole event sequence.
\revised{
    In our experiments, we use GRU, \citep{Cho2014OnTP}, a recurrent network which demonstrates
    robust performance on the sequential data, \citep{Babaev2019ETRNNAD}. In this case,
    $\phi_{\mathrm{seq}}$ is computed by the recurrence $c_{t+1} = \mathrm{GRU}(z_{t+1}, c_t)$
    starting from a learnt $c_0$. We note, that other architectures are possible, including
    LSTM and transformers, \citep{Hochreiter1997LongSM,Vaswani2017AttentionIA}
    (see Section~\ref{sec-ablation}).
}

\medskip
To summarise, CoLES consists of three major ingredients: the event sequence encoder, the positive
and negative pair generation strategy, and the loss function for contrastive learning.


\section{Experiments} \label{sec-exp}

\subsubsection{Datasets}

We compare our method with existing baselines on several publicly available datasets of
event sequences from various data science competitions. We chose datasets with sufficient
amounts of discrete events per user.

\textbf{Age group prediction competition}%
\footnote{
    \url{https://ods.ai/competitions/sberbank-sirius-lesson}
}
The dataset of 44M an\-onymiz\-ed credit card transactions representing 50K individuals
was used to predict the age group of a person. The multiclass target label is known only for 30K records,
and within this subset the labels are balanced. Each transaction includes the date, type,
and amount being charged.


\textbf{Churn prediction competition}%
\footnote{
    \url{https://boosters.pro/championship/rosbank1/}
}
The dataset of 1M an\-onymiz\-ed card transactions representing 10K clients was used to
predict churn probability. Each transaction is characterized by date, type, amount and
Merchant Category Code. The binary target label is known only for 5K clients, and labels
are almost balanced.

\textbf{Assessment prediction competition}%
\footnote{
    \url{https://www.kaggle.com/c/data-science-bowl-2019}
}
The task is to predict the in-game assessment results based on the history of children's
gameplay data. Target is one of four grades, with shares $0.50$, $0.24$, $0.14$, $0.12$.
The dataset consists of 12M gameplay events combined in 330K gameplays representing 18K
children. Only 17.7K gameplays are labeled.
Each gameplay event is characterized by a timestamp, an event code, an incremental counter
of events within a game session, time since the start of the game session, etc.

\textbf{Retail purchase history age group prediction}%
\footnote{
    \url{https://ods.ai/competitions/x5-retailhero-uplift-modeling}
}
The task is to predict the age group of a client based on their retail purchase history.
The age group is known for all clients. The group ratio is balanced in the dataset. The dataset consists
of 45.8M retail purchases representing 400K clients. Each purchase is characterized by
time, product level, segment, amount, value, loyalty program points received.

\revised{
\textbf{Scoring competition}%
\footnote{
     \url{https://boosters.pro/championship/alfabattle2/overview}
}
The dataset of 443M an\-onymiz\-ed credit card transactions representing 1.47M persons was
used to predict the probability of credit product default. The label is known for 0.96M
persons. The default rate is 2.76\% in the dataset. Each
transaction includes the set of date features, set of type features, and amount being
charged.
}


\subsubsection{Repeatability and periodicity of the datasets} \label{sec-period}

\begin{figure*}
  \centering
  \begin{subfigure}{0.25\linewidth}
    \caption{Age group}
    \includegraphics[width=\linewidth]{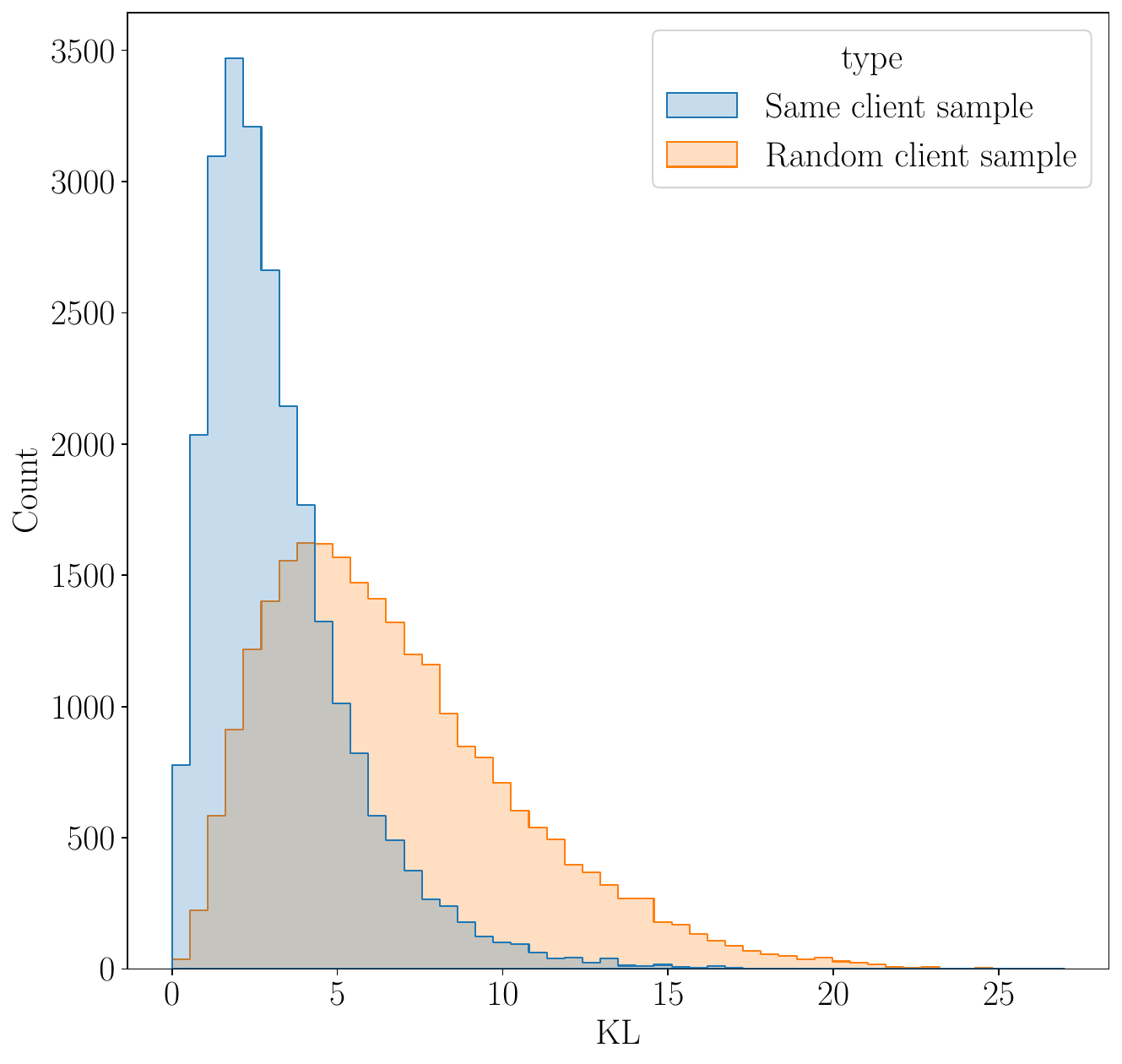}
  \end{subfigure}%
  \begin{subfigure}{0.25\linewidth}
    \caption{Assessment}
    \includegraphics[width=\linewidth]{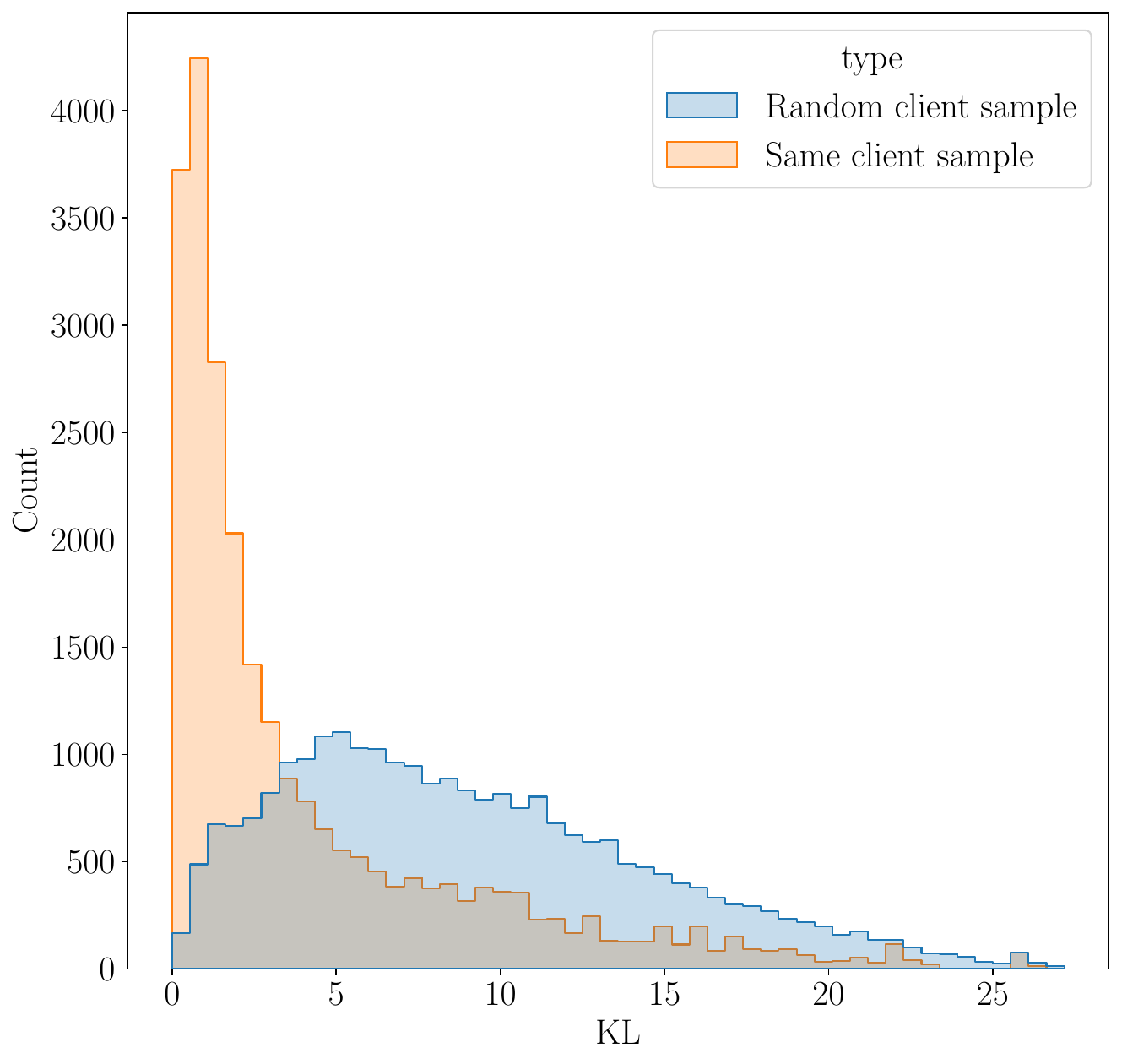}
  \end{subfigure}%
  \begin{subfigure}{0.25\linewidth}
    \caption{Retail}
    \includegraphics[width=\linewidth]{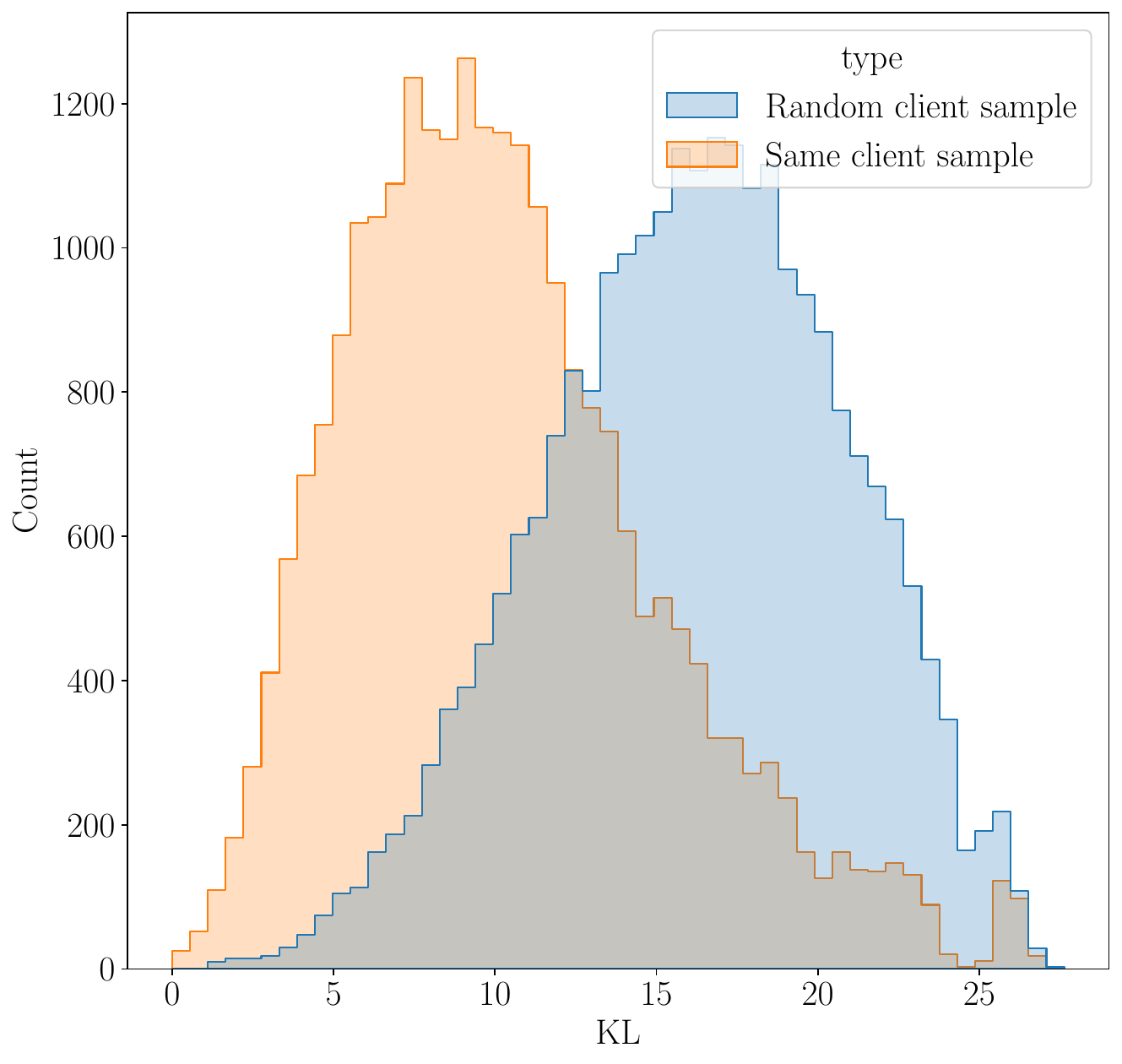}
  \end{subfigure}%
  \begin{subfigure}{0.25\linewidth}
    \caption{Texts}
    \centerline{\includegraphics[width=\linewidth]{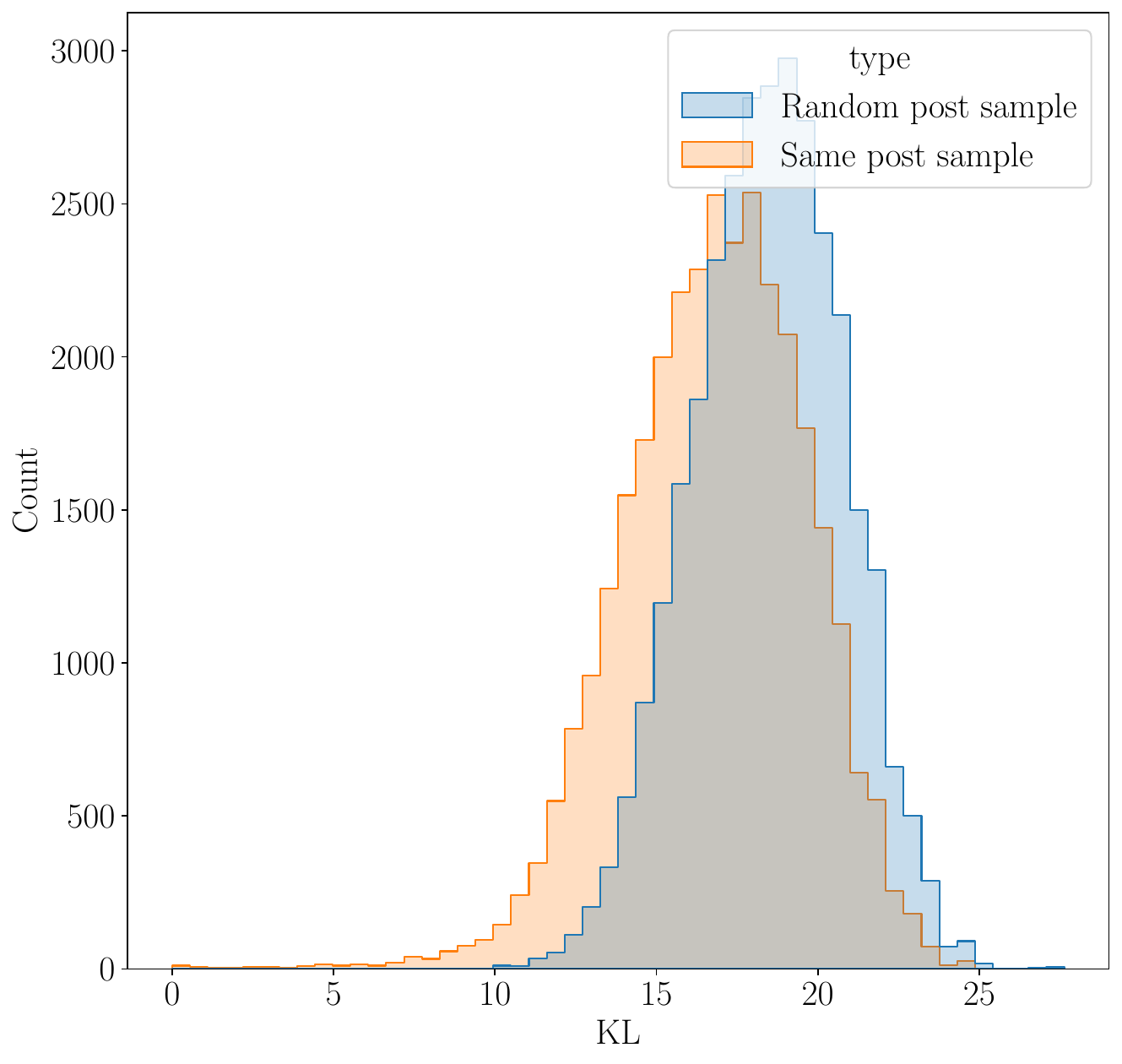}}
    \label{fig-subseq-kl-texts}
  \end{subfigure}
  \caption{
    Periodicity and repeatability of the data. KL-divergence between event types
    of two random sub-sequences from the same sequence is compared with KL-divergence
    between sub-sequences of different sequences. Additional plot (\ref{fig-subseq-kl-texts})
    is provided as an example for data without any repeatable structure.
  }
  \label{fig-subseq-kl}
\end{figure*}

To check that considered datasets follow our repeatability and periodicity assumption
made in Section~\ref{sec-pos-pairs}
we performed the following experiments. We measure the KL-divergence between two kinds
of samples:
(1) between random slices of the same sequence, generated using a modified version of
Algorithm~\ref{alg-slce-ss} where overlapping events are dropped, and
(2) between random sub-samples taken from different sequences. The results are displayed in
Figure~\ref{fig-subseq-kl}, which shows that the KL-divergence between sub-sequences of
the same sequence of events is relatively small compared to the typical KL-divergence between
sub-samples of different sequences of events. This observation supports our repeatability
and periodicity assumption.
Also note that additional plot (\ref{fig-subseq-kl-texts}) is provided as an example for
data without any repeatable structure.

\subsubsection{Dataset split}

For each dataset, we set apart 10\% persons from the labeled part of the data as
the \emph{test set}, used for evaluation. The remaining 90\% of labeled data and
unlabeled data constitute our \emph{training set} used for learning. The hyper-parameters
of each method were selected by random search over $5$-fold cross-validation on
the training set.

For the learning of semi-supervised/self-supervised techniques (including CoLES),
we used all transactions of training sets including unlabeled data. The unlabelled
parts of the datasets were ignored while training supervised models.

\subsubsection{Training performance}

Neural networks were trained on a single Tesla P-100 GPU card. At the training
stage of CoLES, each single training batch is processed in 142 milliseconds. For example,
in the age group prediction dataset a single batch contains 64 unique individuals with
five sub-sequences per each, i.e. 320 training sub-sequences in total, each averaging
90 transactions, which means that each batch contains about 28800 transactions.

\subsubsection{Hyperparameters}

Unless specified otherwise, we use contrastive loss and random slices pair generation strategy
for CoLES in our experiments (see Section~\ref{sec-res} for motivation). The final set of
hyper-parameters used for CoLES is shown in the Table \ref{tab-hyper}.

\begin{table}
    \centering
    \caption{Hyper-parameters for CoLES training}
    \begin{tabularx}{\linewidth}{Xcccc}
        \toprule
            \textbf{Dataset} & \textbf{Age} & \textbf{Churn} & \textbf{Assess} & \textbf{Retail} \\
        \midrule
            \textbf{Embedding size} & 800 & 1024 & 100 & 800 \\
            \textbf{Learning rate} & 0.001 & 0.004 & 0.002 & 0.002 \\
            \textbf{N samples in batch} & 64 & 128 & 256 & 256 \\
            \textbf{N epochs} & 150 & 60 & 100 & 30 \\
            \textbf{Min sequence length} & 25 & 15 & 100 & 30 \\
            \textbf{Max sequence length} & 200 & 150 & 500 & 180 \\
            \textbf{Encoder} & GRU & LSTM & GRU & GRU \\
        \bottomrule
    \end{tabularx}%
    \\
    \small{
        For all methods, a random search on 5-fold cross-validation over the train set is
        used for hyper-parameter selection. The number of sub-sequences generated for each
        sequence was always 5 for each dataset.
    }
    \label{tab-hyper}
\end{table}

\subsection{Baselines} \label{sec-baselines}


\subsubsection{LightGBM}

The Gradient Boosting Machine (GBM)~\citep{Friedman2001GreedyFA} is generally considered a strong
baseline in cases of tabular data with heterogeneous features,%
\citep{Wu2009AdaptingBF,Vorobev2019LearningTS,Zhang2015AGB,Niu2019ACS}.
%
\revised{
In our experiments we use LightGBM~\citep{Ke2017LightGBMAH} as a model for the downstream task
(see Figure~\ref{fig-arch}, Phase 2a) and consider alternative input features:
(1) the vector of of hand-crafted aggregate features produced from the raw transactional data, or
(2) the embedding of the sequence of transactions, produced by the encoder network (see Section~\ref{sec-enc-arch}).
For the latter the encoder model is trained in the self-supervised fashion either with CoLES
or one of its alternatives, described in the Section~\ref{sec-ss-base}.
}

\subsubsection{Hand-crafted features} \label{sec-hand-features}
All attributes of each transaction are either numerical, e.g. amount, or categorical, e.g.
merchant category (MCC code), transaction type, etc.
For the numerical attributes we apply aggregation functions, such as 'sum', 'mean', 'std',
'min', 'max', over all transactions per sequence. For example, if we apply 'sum' over the
numerical field 'amount' we get a feature 'sum of all transaction amounts per sequence'.
Categorical attributes are aggregated in a slightly different way. We group the transactions
separately by every unique value of each categorical attribute and apply an aggregation
function, such as 'count', 'mean', 'std' over all numerical attributes. For example, if
we apply 'mean' for the numerical attribute 'amount' grouped by categorical attribute
'MCC code' we obtain one feature 'mean amount of all transactions for the specific MCC
code' per sequence.

\subsubsection{Self-supervised baselines} \label{sec-ss-base}

We compared CoLES method against major existing approaches to self-supervised embedding, which are
applicable to the event sequence data.

\textbf{NSP.} We consider a simple baseline inspired by the \emph{next sentence prediction} task
in BERT~\citep{Devlin2019BERTPO}. Specifically, we generate two sub-sequences A and B in a way
that 50\% of the time B follows A in the same sequence (positive pair), and 50\% of the time it
is a random fragment from another sequence (negative pair).

\textbf{SOP.} Another baseline is the same as \emph{sequence order prediction} task from
ALBERT~\citep{Lan2020ALBERTAL}, which uses two consecutive sub-sequences as a positive pair,
and two consecutive sub-sequences with swapped order as a negative pair.

\textbf{RTD.} The \emph{replaced token detection} approach from ELECTRA~\citep{Clark2020ELECTRAPT}
could be also adapted to event sequences. To get this baseline, we replaced 15\% of events from
the sequence with random events, taken from other sequences and train a model to predict whether
an event is replaced or not.

\textbf{CPC.} Additionally, we compare against Contrastive Predictive Coding
(CPC)~\citep{Oord2018RepresentationLW} -- a self-supervised learning method that demonstrates state-of-the-art
performance on sequential data audio, computer vision, reinforcement learning and recommender
systems domains,
\citep{Zhou2020ContrastiveLF}.

\subsection{Results} \label{sec-res}

\begin{table}
    \centering
    \caption{Comparison of batch generation strategies}
    \begin{tabularx}{\linewidth}{Xcccc}
        \toprule
            \makecell{\textbf{Sample} \\ \textbf{method}} &
            \makecell{\textbf{Age} \\ \small{Accuracy}} &
            \makecell{\textbf{Churn} \\ \small{AUROC}} &
            \makecell{\textbf{Assess} \\ \small{Accuracy}} &
            \makecell{\textbf{Retail} \\ \small{Accuracy}} \\
        \midrule
            \textbf{Random samples} & 0.613 & 0.820 & 0.563 & 0.523 \\
            \textbf{Random disjoint samples} & 0.619 & 0.819 & 0.563 & 0.505 \\
            \textbf{Random slices} & \textbf{0.639} & \textbf{0.823} & \textbf{0.618} & \textbf{0.542} \\
        \bottomrule
    \end{tabularx}%
    \\
    \small{5-fold cross-validation metric is shown}
    \label{tab-pair-gen}
\end{table}

\subsubsection{Discussion of design choices} \label{sec-ablation}

To evaluate the proposed method of sub-sequence generation (see Section~\ref{sec-pos-pairs})
we compared it with two alternatives: \revised{
    (1) The random sampling without replacement strategy, similar to
    \citep{Yao2020SelfsupervisedLF}, and
    (2) random disjoint samples strategy, which resembles the generation
    proposed in \citep{Ma2020DisentangledSI}.
    The first approach generates a non-contiguous sub-sequence of events, by
    repeatedly drawing a random event from the sequence \emph{without replacement}
    preserving the in-sequence order of the sampled events.
    The second approach produces sub-sequences by the randomly splitting
    the initial sequence into several non-overlapping contiguous segments.
    The motivation is that overlaps between sub-sequences may possibly lead to
    overfitting, since the exact sub-sequences of events are the same and may be
    ``memoized'' by the encoder without learning underlying similarities.
}

The results of the comparison are presented in Table~\ref{tab-pair-gen}. The proposed strategy
of generating random sub-sequence slices consistently outperforms alternatives.

\revised{
    We evaluated several contrastive learning loss functions that showed promising
    performance on different datasets~\citep{Kaya2019DeepML} and some classical variants, namely: 
    contrastive~loss~\citep{Hadsell2006DimensionalityRB}, binomial deviance loss~\citep{Yi2014DeepML},
    triplet loss \citep{Hoffer2015DeepML}, histogram~loss~\citep{Ustinova2016LearningDE}, and
    margin~loss~\citep{Manmatha2017SamplingMI}. The results of comparison are shown in Table~\ref{tab-loss-type}.
    Noteworthy, although the contrastive loss can be considered as the basic variant of
    contrastive learning, yet it still manages to achieve strong results on the downstream tasks.
    We speculate that an increase in the model's performance on contrastive learning task does
    not always lead to increased performance on downstream tasks.
}

We also compared popular negative sampling strategies (distance-weighted sampling~\citep{Manmatha2017SamplingMI},
and hard-negative mining~\citep{Schroff2015FaceNetAU}) with random negative sampling strategy (see
Table \ref{tab-neg-sampl}).
We can observe that hard negative mining leads to a measurable increase in quality on downstream
tasks in comparison to random negative sampling.

\revised{
Yet another possible design choice of the method is the encoder architecture. We compared
several popular options for the sequence encoder: GRU~\citep{Cho2014OnTP}, LSTM~\citep{Hochreiter1997LongSM} and Transformer~\citep{Vaswani2017AttentionIA}. Table~\ref{tab-enc-type} shows that the choice of
the encoder architecture has little effect on the performance of the proposed method.
}

\begin{table}
    \centering
    \caption{Comparison of encoder types}
    \begin{tabular}{lcccc}
        \toprule
        \textbf{Dataset} &
            \makecell{\textbf{Age group} \\ \small{Accuracy}} &
            \makecell{\textbf{Churn} \\ \small{AUROC}} &
            \makecell{\textbf{Assess} \\ \small{Accuracy}} &
            \makecell{\textbf{Retail} \\ \small{Accuracy}} \\
        \midrule
            \textbf{LSTM} & 0.621 & \textbf{0.823} & \textbf{0.620} & 0.535 \\
            \textbf{GRU} & \textbf{0.638} & 0.812 & 0.618 & \textbf{0.542} \\
            \textbf{Transformer} & 0.622 & 0.780 & 0.542 & 0.499 \\
        \bottomrule
    \end{tabular}%
    \\
    \small{5-fold cross-validation metric is shown}
    \label{tab-enc-type}
\end{table}

\begin{table}
    \centering
    \caption{Comparison of contrastive learning losses}
    \begin{tabularx}{\linewidth}{Xcccc}
        \toprule
        \textbf{Dataset} &
            \makecell{\textbf{Age group} \\ \small{Accuracy}} &
            \makecell{\textbf{Churn} \\ \small{AUROC}} &
            \makecell{\textbf{Assess} \\ \small{Accuracy}} &
            \makecell{\textbf{Retail} \\ \small{Accuracy}} \\
        \midrule
            \textbf{Contrastive} \small{margin=0.5} & \textbf{0.639} & \textbf{0.823} & \textbf{0.618} & \textbf{0.542} \\
            \textbf{Binomial deviance} & 0.621 & 0.769 & 0.589 & 0.535 \\
            \textbf{Histogram} & 0.632 & 0.815 & 0.615 & 0.533 \\
            \textbf{Margin} & 0.638 & \textbf{0.823} & 0.612 & 0.541 \\
            \textbf{Triplet} & 0.636 & 0.781 & 0.600 & 0.541 \\
        \bottomrule
    \end{tabularx}%
    \\
    \small{5-fold cross-validation metric is shown}
    \label{tab-loss-type}
\end{table}

\begin{table}
    \centering
    \caption{Comparison of negative sampling strategies}
    \begin{tabularx}{\linewidth}{Xcccc}
        \toprule
        \textbf{Dataset} &
            \makecell{\textbf{Age group} \\ \small{Accuracy}} &
            \makecell{\textbf{Churn} \\ \small{AUROC}} &
            \makecell{\textbf{Assessment} \\ \small{Accuracy}} &
            \makecell{\textbf{Retail} \\ \small{Accuracy}} \\
        \midrule
            \textbf{Hard negative mining} & \textbf{0.639} & \textbf{0.823} & \textbf{0.618} & \textbf{0.542} \\
            \textbf{Random negative sampling} & $0.626$ & $0.815$ & $0.593$ & $0.530$ \\
            \textbf{Distance-weighted sampling} & $0.629$ & $0.821$ & $0.603$ & $0.536$ \\
        \bottomrule
    \end{tabularx}%
    \\
    \small{5-fold cross-validation metric is shown}
    \label{tab-neg-sampl}
\end{table}

\begin{figure*}
  \centering
  \begin{subfigure}{0.25\linewidth}
    \caption{Age group}
    \includegraphics[width=\linewidth]{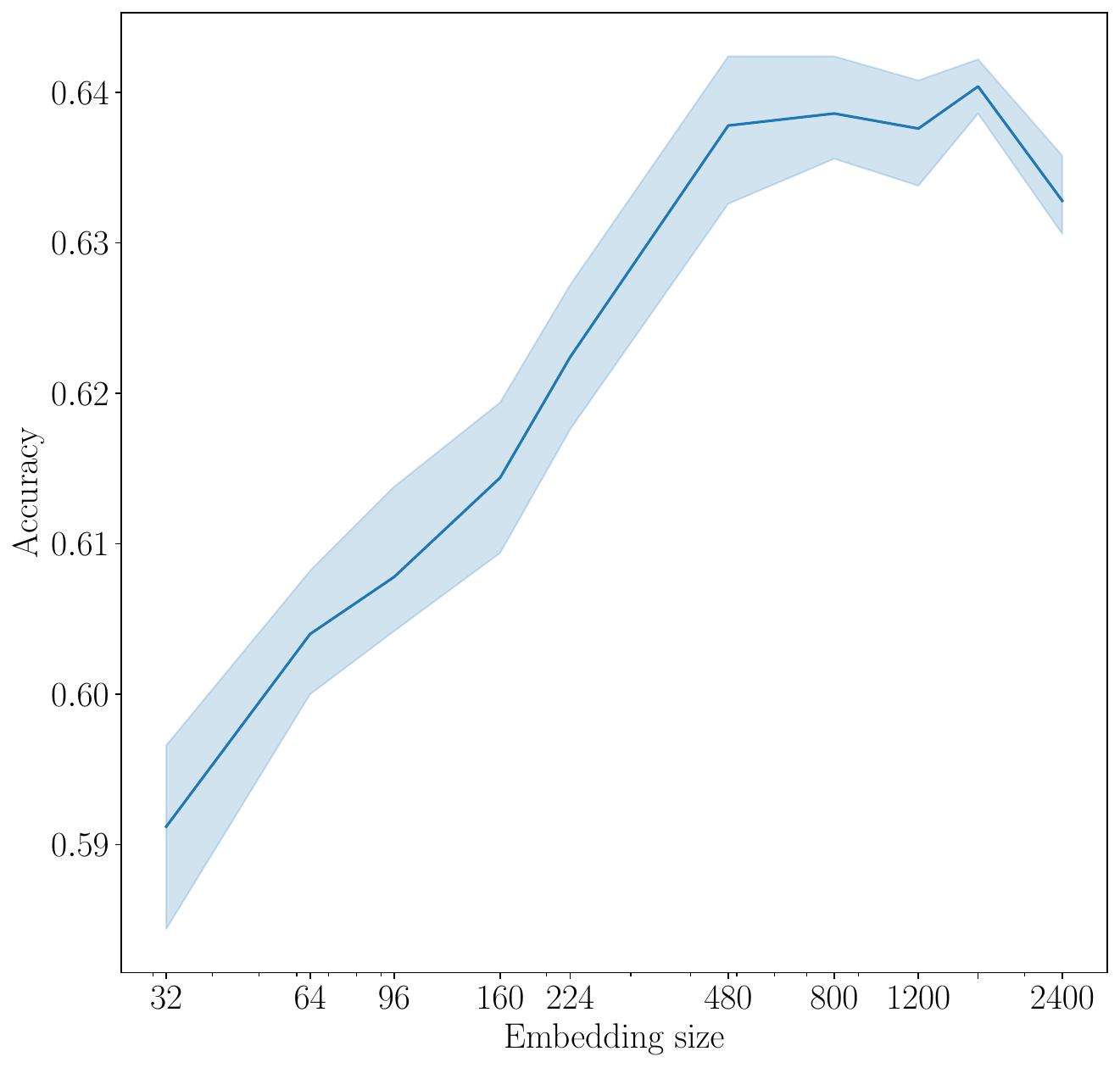}
  \end{subfigure}%
  \begin{subfigure}{0.25\linewidth}
    \caption{Churn}
    \includegraphics[width=\linewidth]{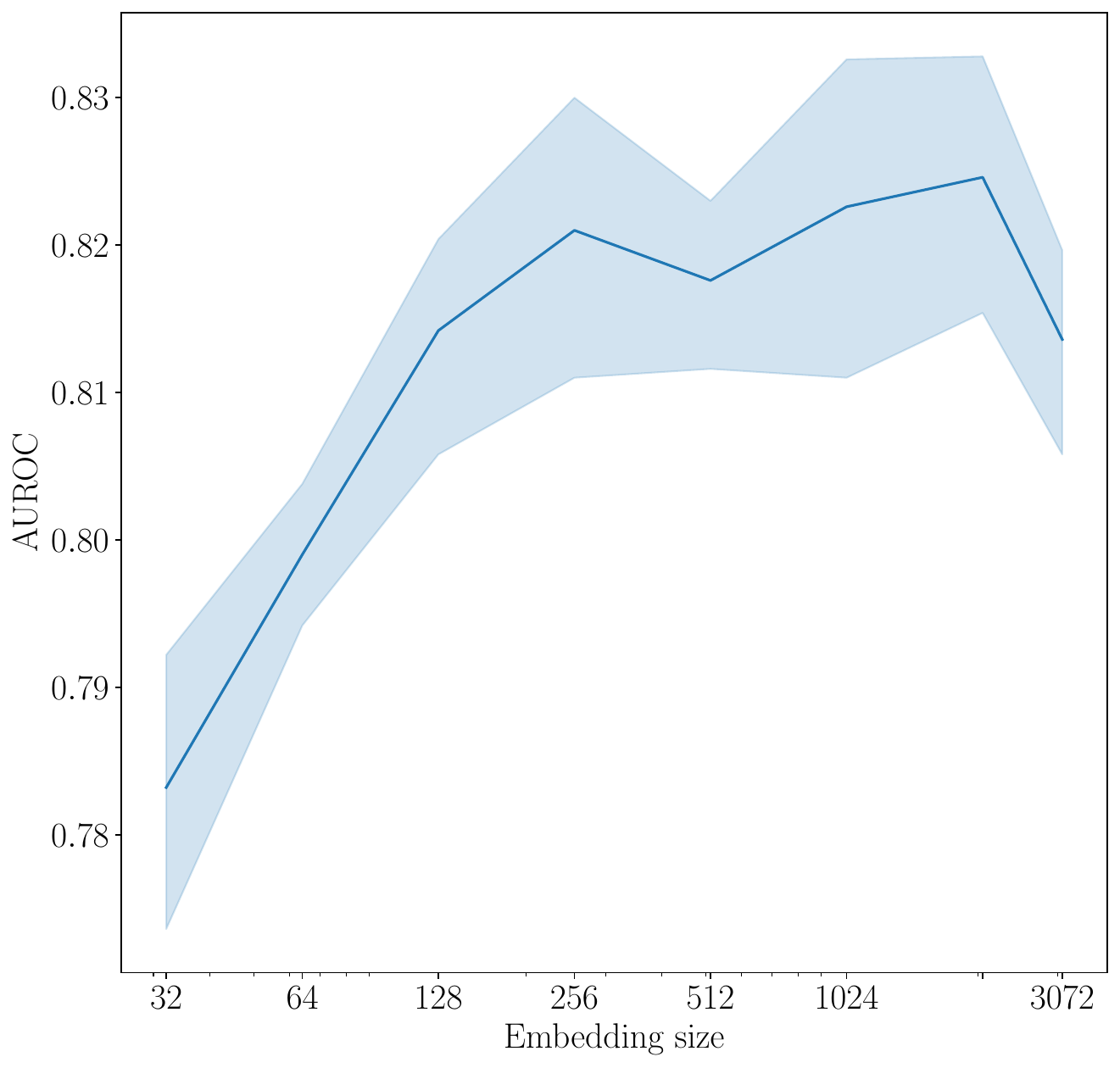}
  \end{subfigure}%
  \begin{subfigure}{0.25\linewidth}
    \caption{Assessment}
    \includegraphics[width=\linewidth]{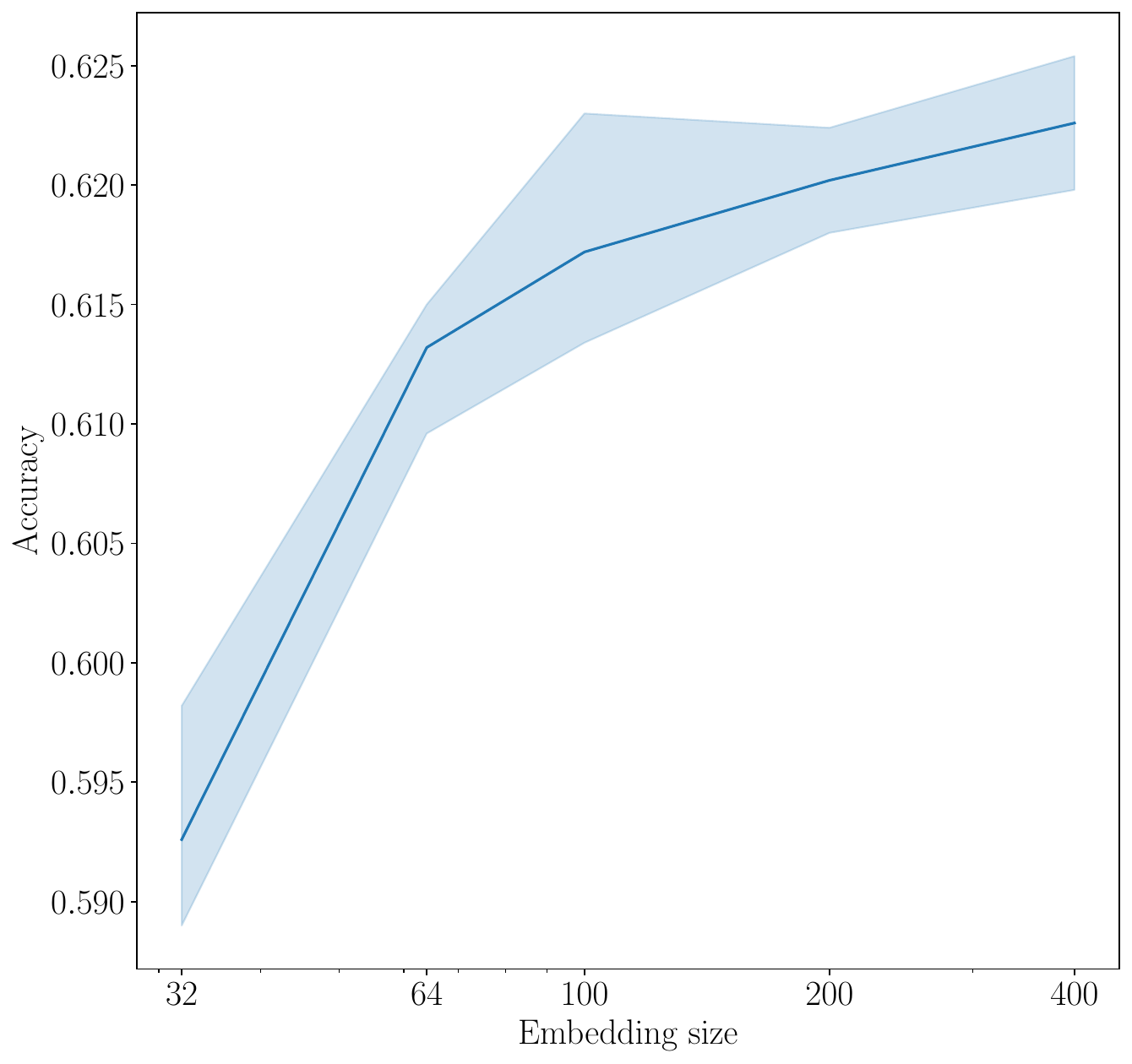}
  \end{subfigure}%
  \begin{subfigure}{0.25\linewidth}
    \caption{Retail}
    \includegraphics[width=\linewidth]{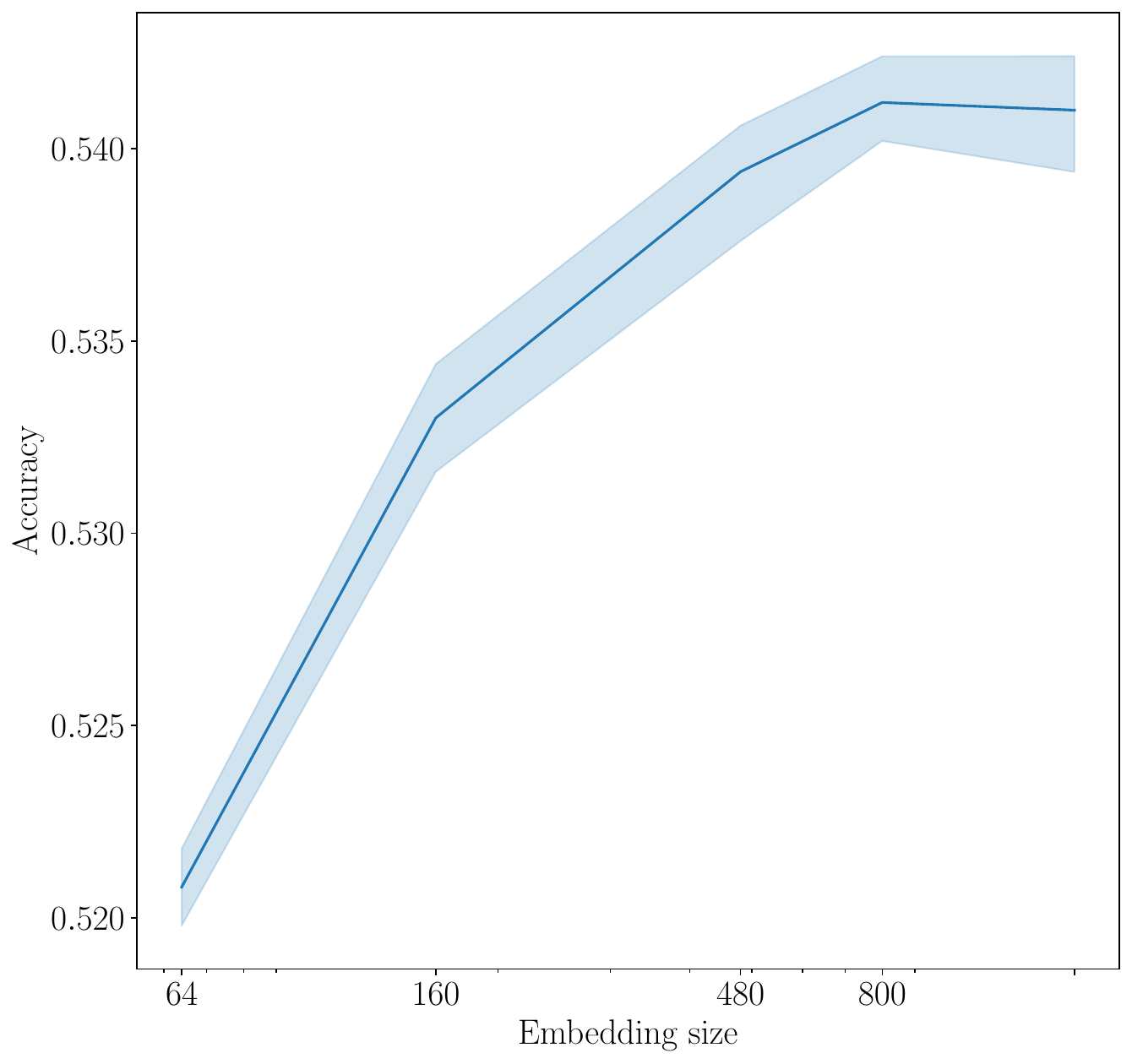}
  \end{subfigure}
  \caption{Embedding dimensionality vs. quality}
  \label{fig-emb-dim}
\end{figure*}

\revised{
Figure \ref{fig-emb-dim} shows that the performance on the downstream task exhibits diminishing
gains as the dimensionality of the embedding increases. These results can be interpreted through
the lens of the bias-variance trade-off: when the dimensionality is too low, too much information
is discarded (high bias), however, when it is is too high, then much more irrelevant noise seeps
into the embedding (high variance).
Note, that the training time and the memory consumption scale linearly with the embedding size.
}

\subsubsection{Self-supervised embeddings}

We compared CoLES with baselines described in Section~\ref{sec-baselines} in two scenarios.
First, we compared embeddings produced by CoLES with other types of embeddings, including
the manually created ones, by using them as input into a downstream LightGBM model
(see Figure~\ref{fig-arch}, Phase 2a), trained independently from the sequence encoder.
As Table~\ref{tab-downstream-res-emb} demonstrates, the sequence embeddings, produced
by CoLES, perform on par and sometimes even better than manually engineered features.
Furthermore, CoLES consistenlty outperforms other self-supervised baselines on \revised{
    the four out of the five considered datasets
}.

\revised{
    On the scoring dataset, which is larger than the other datasets, CoLES, CPC, and RTD, outperform the hand-crafted baseline.
    The autoregressive nature of the CPC aligns well with the credit scoring task, for which
    the more recent the information the more relevant it is to the target. The RTD embeddings
    are also effective for credit scoring, since the method relies on anomaly detection.

    Simple hand-crafted features can achieve competitive results if the events have a clear
    structure for designing them, e.g. it is straightforward to compute the group-wise aggregate
    statistics of historical data based on some attribute. In the commercial settings (Section~\ref{sec-commercial})
    the situation may be different: it is non-trivial to manufacture meaningful features for
    transactions of legal entities, since it is not clear what the natural groupings are. We
    discuss the difference between simple events (card transactions) and more complex events
    (transactions of the legal entities) at the end of Section~\ref{sec-commercial}.
}

\begin{table*}
    \centering
    \caption{Quality of unsupervised embeddings as features for the downstream task}
    \begin{tabularx}{\linewidth}{Xccccc}
    \toprule
        \textbf{Method} &
        \makecell{\textbf{Age} \\ \small{Accuracy}} &
        \makecell{\textbf{Churn} \\ \small{AUROC}} &
        \makecell{\textbf{Assess} \\ \small{Accuracy}} &
        \makecell{\textbf{Retail} \\ \small{Accuracy}} &
        \makecell{\revised{\textbf{Scoring}} \\ \small{AUROC}}\\
    \midrule
        \textbf{Designed features} & $0.631$\tiny{$\pm 0.003$} & $0.825$\tiny{$\pm 0.004$} & \bm{$0.602$}\tiny{$\pm 0.005$} & \bm{$0.547$}\tiny{$\pm 0.001$} & $0.779$\tiny{$\pm 0.001$} \\
        \textbf{SOP} & $0.493$\tiny{$\pm 0.002$} & $0.782$\tiny{$\pm 0.005$} & $0.577$\tiny{$\pm 0.002$} & $0.428$\tiny{$\pm 0.001$} & $0.724$\tiny{$\pm 0.001$} \\
        \textbf{NSP} & $0.622$\tiny{$\pm 0.004$} & $0.830$\tiny{$\pm 0.004$} & $0.581$\tiny{$\pm 0.003$} & $0.425$\tiny{$\pm 0.002$} & $0.766$\tiny{$\pm 0.001$} \\
        \textbf{RTD} & $0.632$\tiny{$\pm 0.002$} & $0.801$\tiny{$\pm 0.004$} & $0.580$\tiny{$\pm 0.003$} & $0.520$\tiny{$\pm 0.001$} & $0.791$\tiny{$\pm 0.001$} \\
        \textbf{CPC} & $0.594$\tiny{$\pm 0.002$} & $0.802$\tiny{$\pm 0.003$} & $0.588$\tiny{$\pm 0.002$} & $0.525$\tiny{$\pm 0.001$} & $0.791$\tiny{$\pm 0.001$} \\
        \textbf{CoLES} & \bm{$0.638$}\tiny{$\pm 0.007$} & \bm{$0.843$}\tiny{$\pm 0.003$} & $0.601$\tiny{$\pm 0.002$} & $0.539$\tiny{$\pm 0.001$} & \bm{$0.792$}\tiny{$\pm 0.001$} \\
    \bottomrule
    \end{tabularx}%
    \\
    \small{average test set quality metric and its standard deviation for 5 runs on different folds is shown}
    \label{tab-downstream-res-emb}
\end{table*}

\subsubsection{Fine-tuned embeddings}

In the second scenario, we fine-tune pre-trained models for specific downstream tasks (see
Figure~\ref{fig-arch}, Phase 2b). The models are pre-trained using CoLES or another self-supervised
learning approaches and then are trained on the labeled data for the specific end task.

\revised{
    The fine-tuning step is done by adding a classification network $h$ (single-layer neural
    network with softmax activation) to the pre-trained encoder network $M$ (see Section~\ref{sec-enc-arch}).
    Both networks are trained jointly on the downstream task, i.e. the classifier takes the output
    of the encoder and produces a prediction ($\hat{y} = h(M(\{x\}))$) and its error propagates
    back through both.
}
In addition to the aforementioned baselines, we compare our method to a supervised learning
approach, where the encoder network $M$ is not pre-trained using self-supervised target.

As Table~\ref{tab-downstream-res} shows, CoLES representations obtained after fine-tuning achieve
superior performance on all the considered datasets, outperforming other methods by significant
margins.

\begin{table}
    \centering
    \caption{Quality of the pre-trained model on the downstream tasks}
    \begin{tabularx}{\linewidth}{Xcccc}
        \toprule
            \textbf{Method}
            & \makecell{\textbf{Age} \\ \small{Accuracy}}
            & \makecell{\textbf{Churn} \\ \small{AUROC}}
            & \makecell{\textbf{Assess} \\ \small{Accuracy}}
            & \makecell{\textbf{Retail} \\ \small{Accuracy}}
            \\
        \midrule
            \textbf{Designed features} & $0.631$\tiny{$\pm 0.003$} & \bm{$0.825$}\tiny{$\pm 0.004$} & $0.602$\tiny{$\pm 0.005$} & $0.547$\tiny{$\pm 0.001$} \\
            \textbf{Super\-vised learning} & $0.628$\tiny{$\pm 0.004$} & $0.817$\tiny{$\pm 0.009$} & $0.602$\tiny{$\pm 0.005$} & $0.542$\tiny{$\pm 0.001$}\\
            \textbf{RTD pretrain} & $0.635$\tiny{$\pm 0.006$} &  $0.819$\tiny{$\pm 0.005$} & $0.586$\tiny{$\pm 0.003$} & $0.544$\tiny{$\pm 0.002$} \\
            \textbf{CPC pretrain} & $0.615$\tiny{$\pm 0.009$} &  $0.810$\tiny{$\pm 0.006$} & $0.606$\tiny{$\pm 0.004$} & $0.549$\tiny{$\pm 0.001$} \\
            \textbf{CoLES pretrain} & \bm{$0.644$}\tiny{$\pm 0.004$} & \bm{$0.827$}\tiny{$\pm 0.004$} & \bm{$0.615$}\tiny{$\pm 0.003$} & \bm{$0.552$}\tiny{$\pm 0.001$} \\
        \bottomrule
    \end{tabularx}%
    \\
    \small{average test set quality metric and its standard deviation for 5 runs on different folds is shown}
    \label{tab-downstream-res}
\end{table}

\subsubsection{Semi-supervised setup}

\begin{figure*}
  \centering
  \begin{subfigure}{0.25\linewidth}
    \caption{Age group}
    \includegraphics[width=\linewidth]{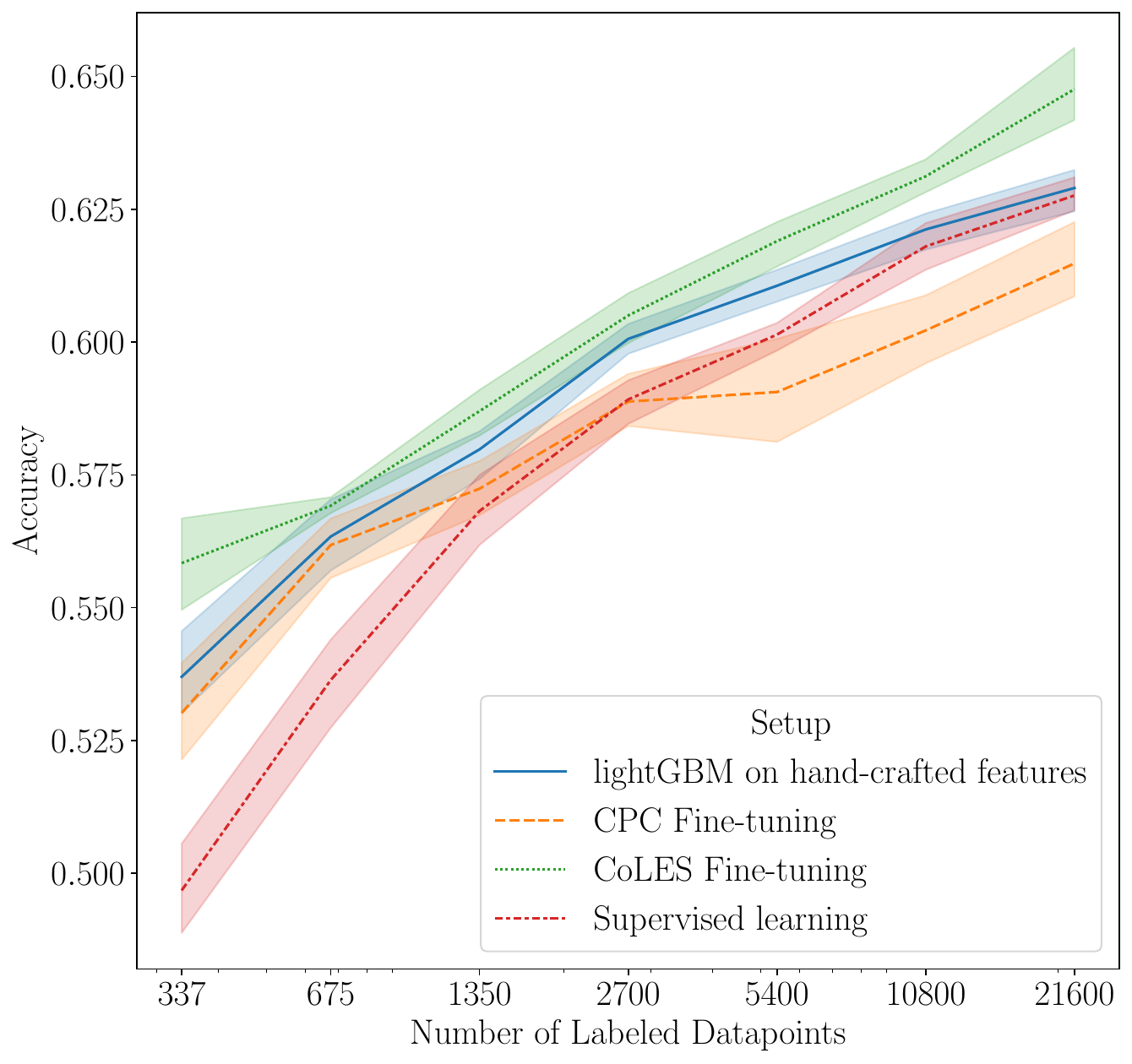}
  \end{subfigure}%
  \begin{subfigure}{0.25\linewidth}
    \caption{Churn}
    \includegraphics[width=\linewidth]{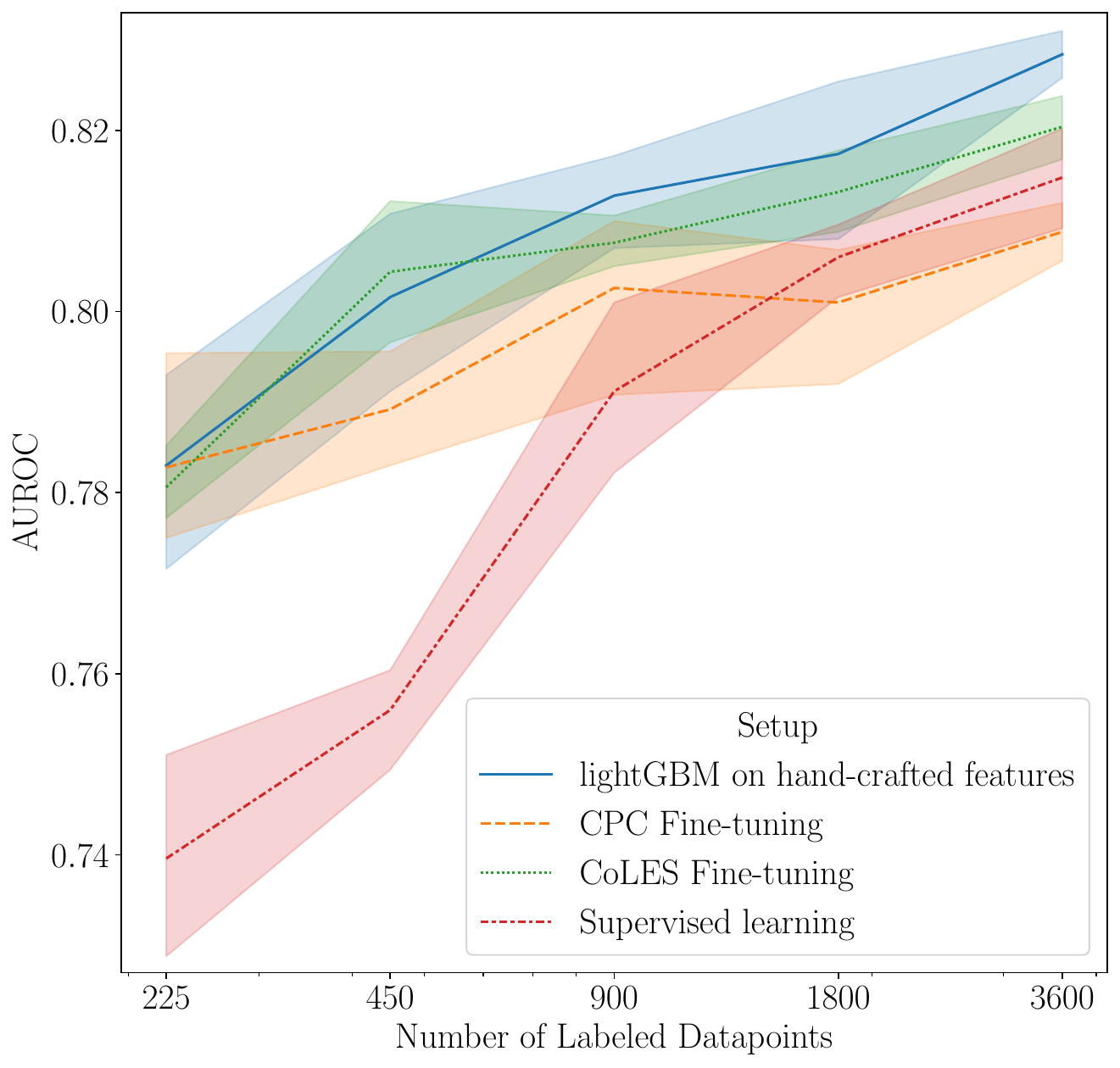}
  \end{subfigure}%
  \begin{subfigure}{0.25\linewidth}
    \caption{Assessment}
    \includegraphics[width=\linewidth]{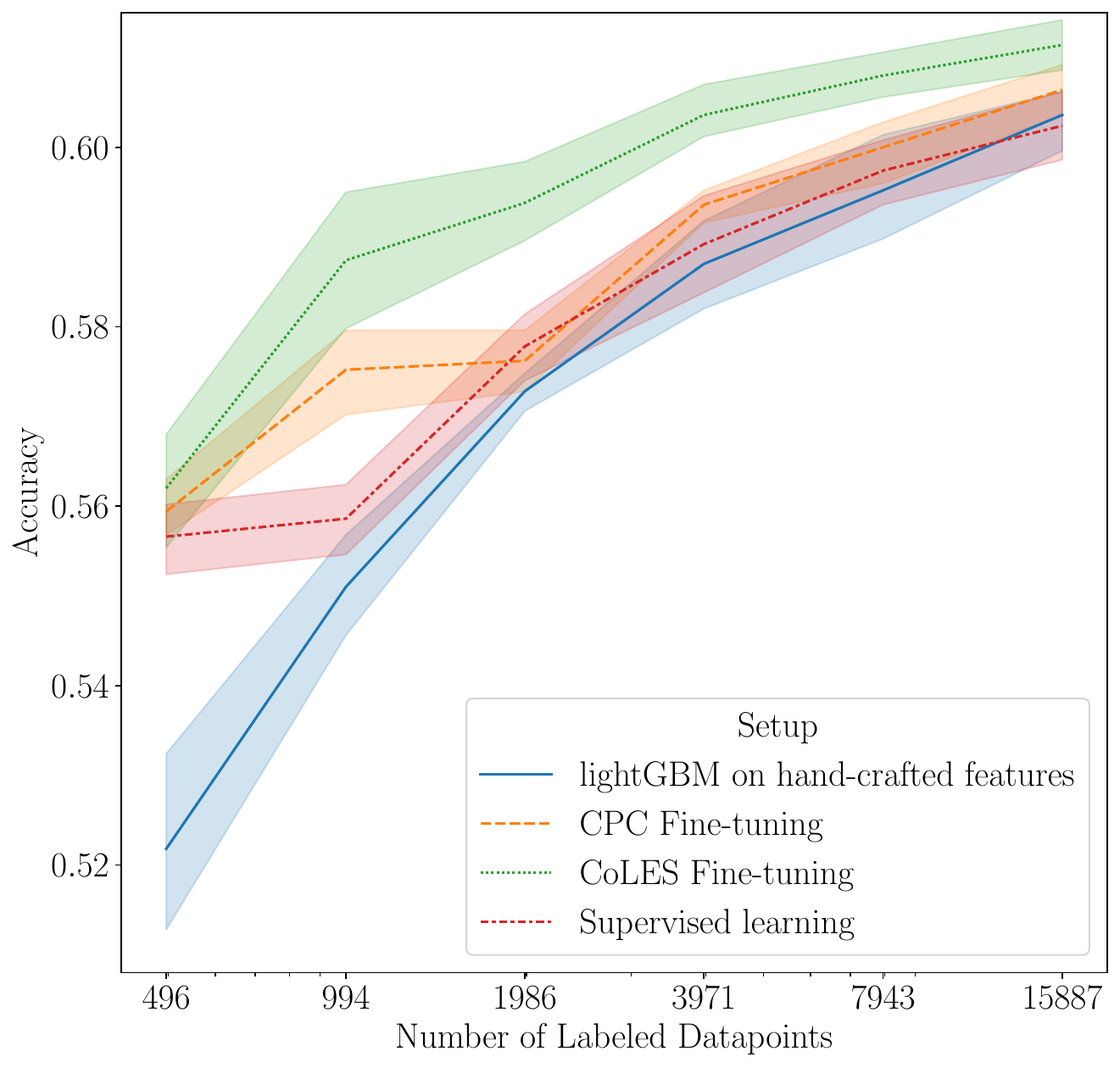}
  \end{subfigure}%
  \begin{subfigure}{0.25\linewidth}
    \caption{Retail}
    \includegraphics[width=\linewidth]{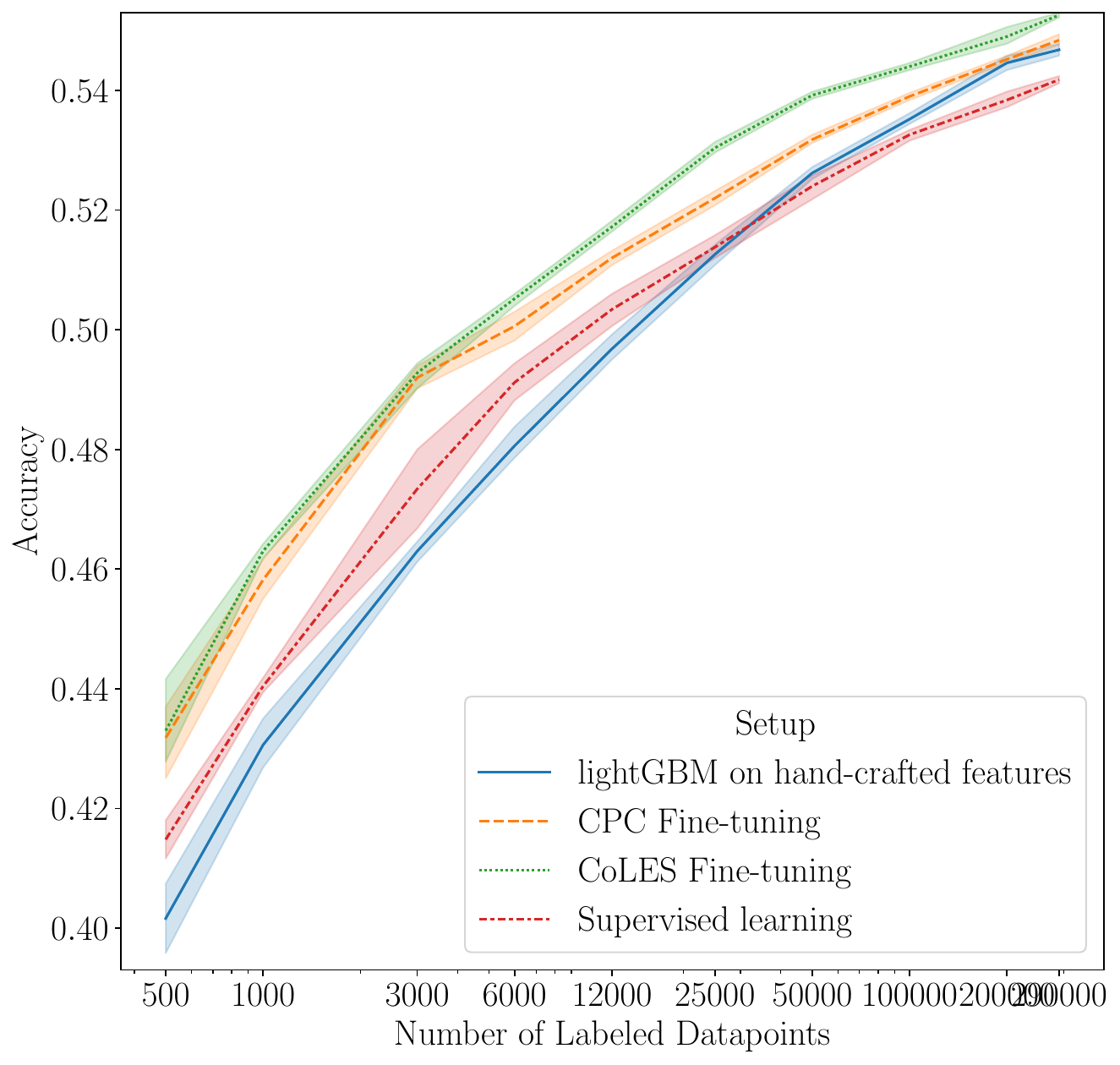}
  \end{subfigure}
  \caption{
    Model quality for different dataset sizes
  }
  \small{The rightmost point corresponds to all labels and supervised setup.}
  \label{fig-semi-main}
\end{figure*}

Here we study the applicability of our method in scenarios where the amount of labeled examples
is limited. We performed a series of experiments where only a random fraction of available labels
is used to train the downstream task models. As in the case of the supervised setup, we compare
the proposed method with hand-crafted features, CPC, and supervised learning without pre-training.
The results are presented in Figure \ref{fig-semi-main}. Note that the performance improvement of
CoLES in comparison to the supervised-only methods increases as we decrease the portion of labeled
examples in the training dataset. Also note that CoLES consistently outperforms CPC for different
volumes of labeled data.

\subsection{CoLES Embeddings in Commercial Settings} \label{sec-commercial}

\begin{table}
    \centering
    \caption{Data structure of the credit card transactions}
    \begin{tabular}{llllll}
        \toprule
            \textbf{Date}
            & \textbf{Amount}
            & \textbf{Currency}
            & \textbf{Country}
            & \makecell{\textbf{Merchant} \\ \textbf{Type}}
            \\
        \midrule
            Jun 21 & 230 & EUR & France & Restaurant \\
            Jun 21 & 5 & USD & US & Transportation \\
            Jun 22 & 40 & USD & US & \makecell[l]{Household \\ Appliance} \\
        \bottomrule
    \end{tabular}
    \label{tab-tr-data}
\end{table}

\begin{table}
    \centering
    \caption{Data structure of the money transfers between legal entities}
    \begin{tabular}{lllllll}
        \toprule
            \textbf{Date}
            & \textbf{Amount}
            & \textbf{Currency}
            & \textbf{Sender}
            & \textbf{Receiver}
            & \textbf{Type}
            \\
        \midrule
            Jul 11 & 20000 & EUR & 1232323 & 6345433 & 23 \\
            Jul 11 & 5000 & USD & 5424443 & 1232323 & 12 \\
            Jul 12 & 14000 & USD & 1232323 & 5424443 & 14 \\
        \bottomrule
    \end{tabular}
    \small{
        The information about company business and the company region is encoded
        in the first letters of the company identifier stored in ``Sender'' and
        ``Receiver'' fields.
    }
\label{tab-org-tr-data}
\end{table}

We applied the proposed self-supervised CoLES method to several machine learning tasks, routinely
considered in a large European financial services company.
In particular, two types embeddings were created:
(1) \emph{legal entity embeddings} of the small- and medium-size companies, based on commercial
transactions and operational histories of their businesses, and
(2) \emph{individual embeddings} of individual/retail customers, 
based on their debit/credit card transaction histories.
Tables~\ref{tab-tr-data} and~\ref{tab-org-tr-data} provide examples of the transactional data
used for building individuals' and legal entities' embeddings. Overall, the dataset of ten
million corporate clients with on average 200 transactions per client was used to train the
embeddings of type (1), and the dataset of five million individual clients with the mean
number of 400 transactions per client was used to train the model for embeddings of type (2).
These two ``in-house'' commercial datasets are considerable larger than the public datasets
outlined in Section~\ref{sec-exp}, and the extra volume of self-supervised training data
allowed us to generate embeddings of significantly higher quality than on the publicly
available data.

We performed extensive evaluation of CoLES embeddings on these in-house datasets by applying
them to different downstream tasks. Legal entity embeddings were applied in the following
use cases:
\begin{itemize}
    \item \textbf{Corporate medical insurance lead generation.}
    In this task, the model should be able to predict client's interest in a corporate medical insurance product.

    \item \textbf{Credit lead generation} for small and medium size businesses.
    In this task, the model should predict an interest of a company in taking a credit.

    \item \textbf{Credit scoring} for small and medium size businesses.
    In this task, the model should predict the the probability of a company's default.

    \item \textbf{Holding structure restoration.}
    In this task, the model should predict if a pair of companies are in the same holding of companies.

    \item \textbf{Fraudulent money transfers monitoring.}
    In this task, the model is used to estimate the likelihood that a particular transaction is fraudulent.
\end{itemize}
As opposed to legal entities, individual embeddings are employed in following downstream tasks:
\begin{itemize}
    \item \textbf{Retail credit scoring.}
    In this task, the model should estimate the probability of default when a client is taking a retail credit.

    \item \textbf{Customer churn prediction.}
    In this task, the model should predict the possibility that a client would stop using the company
    products (cards and deposit accounts).

    \item \textbf{Life insurance lead generation.}
    In this task, the model should predict an interest in a life insurance product.
\end{itemize}

We considered the following three scenarios in most of these tasks:
(1) the baseline scenario -- only the hand-crafted features were used;
(2) the CoLES scenario -- the produced self-supervised embeddings serve as features;
(3) the hybrid (Baseline + CoLES) scenario, combining the hand-crafted features and CoLES embeddings.
In all of these three scenarios, we used the LightGBM method~\citep{Ke2017LightGBMAH} for modelling
in the downstream task. In the first and the third scenarios, we deployed the sets of hand-crafted
features, previously utilized in the organization (see Section~\ref{sec-hand-features} for examples
of the used hand-crafted features).
%

The results of our experiments are presented in Tables~\ref{tab-internal-company} and~\ref{tab-internal-person}.
We observe that CoLES embeddings significantly improve upon the hand-crafted features in terms
of the test performance on end task.

Note that it is more difficult to design valuable hand-crafted features for legal entities than
for individual customers. A typical feature is some statistic aggregated over groups of transactions
on some level. For example, one can aggregate card transactions of an individual customer on
the level of their ``merchant type'' (MCC) field. In contrast, it is unclear how to group fund
transfers of a company by the ``receiver'' field (see examples in Table~\ref{tab-org-tr-data}).
It is hard to manually find a perfect level of aggregation for receivers, since they can be grouped
in many different ways, e.g. by region, size, the type of business, etc. We believe that CoLES is
able to automatically learn a suitable aggregation level. This is one of the reasons why in our
experiments the legal entity embeddings demonstrate higher relative improvements with respect to
hand-crafted features than individual embeddings.

\subsubsection{Deployment details} \label{sec-deployment}

\revised{
    In production environments, our method is applied in two stages:
    training of the encoder neural net $M$ (the training part), followed by calculation of
    the embeddings with it (the inference part).
    We used only a part of the available data for training (10 million corporate clients, and 5 million
    individual clients), but applied the learnt embedder to all available transactional data,
    with more than 90 million cards in total.
    We did not use any of the available distributed training techniques~\citep{Recht2011HogwildAL, Zhao2016FastAP, Kennedy2019APA} during the training part.
    During the inference state we leveraged horizontal scaling scheme, wherein different
    sequences are processed independently in parallel on different nodes of a Hadoop cluster.
}

In order to minimise the efforts of deploying CoLES embeddings inside the company, we used
an ETL process for incremental recalculation of embeddings upon arrival of new transactional data.
Specifically, unlike transformers, recurrent encoders $\phi_{\mathrm{enc}}$, such as
GRU~\citep{Cho2014LearningPR}, reuse prior computations and enable incremental calculation:
the embedding $c_{t + k}$ can be computed iteratively from $c_t$ and $(z_{t+j})_{j=1}^k$,
using $c_{t + j} = \phi_{\mathrm{enc}}(z_{t + j}, c_{t+j-1})$. This architectural choice
reduces the inference time needed for updating the embeddings online. 

Furthermore, by employing a quantization technique it is possible to compress
the sequence embeddings, without much loss in performance on downstream tasks.
%
For instance, single precision values in the embedding could be mapped into
the range from 0 to 15, which makes a 256-dimensional embedding which originally
took 1Kb, take only 128 bytes.
%
%

\begin{table}
    \centering
    \caption{
        Performance comparison of CoLES-based models with the baselines across downstream
        tasks for the legal entities
    }
    \begin{tabularx}{\linewidth}{Xccc}
        \toprule
            \textbf{Task}
            & \textbf{Baseline }
            & \textbf{CoLES}
            & \makecell{\textbf{Baseline} \\ \textbf{+ CoLES}}
            \\ 
        \midrule
            \textbf{Insurance lead generation} & 0.71 & 0.85 & \textbf{0.85} \\
            \textbf{Credit lead generation} & 0.75 & 0.79 & \textbf{0.79} \\
            \textbf{Credit scoring} & 0.73 & 0.71 & \textbf{0.77} \\
            \textbf{Holding structure restoration} & 0.92 & 0.97 & \textbf{0.97} \\
            \textbf{Fraud monitoring} & 0.82 & 0.84 & \textbf{0.85} \\
        \bottomrule
    \end{tabularx}%
    \\
    \small{
        Baseline includes both transactional and non-transactional hand-crafted
        features. AUROC is used as quality metric.
    }
    \label{tab-internal-company}
\end{table}

\begin{table}
    \centering
    \caption{
        Performance comparison of CoLES-based models with the baselines across downstream
        tasks for the retail customers
    }
    \begin{tabularx}{\linewidth}{Xccc}
        \toprule
            \textbf{Task }
            & \textbf{Baseline}
            & \textbf{CoLES}
            & \makecell{\textbf{Baseline}\\\textbf{+ CoLES}}
            \\
        \midrule
            \textbf{Retail credit scoring} & 0.88 & 0.87 & \textbf{0.92} \\
            \textbf{Customer churn} & 0.74 & 0.65 & \textbf{0.76} \\
            \textbf{Insurance lead generation} & 0.75 & 0.74 & \textbf{0.78} \\ 
        \bottomrule
    \end{tabularx}%
    \\
    \small{
        Baseline includes both transactional and non-transactional hand-crafted
        features. AUROC is used as quality metric.
    }
    \label{tab-internal-person}
\end{table}

\section{Conclusions} \label{sec-conclusions}

In this paper, we present \emph{Contrastive Learning for Event Sequences (CoLES)}, a novel
self-supervised method for building embeddings of discrete event sequences. CoLES can be
efficiently used to produce embeddings of complex event sequences for various downstream
tasks.

We empirically demonstrate that our approach achieves strong performance results on
several downstream tasks and consistently outperforms both classical machine learning
baselines on hand-crafted features, as well as on several existing self-supervised and
semi-supervised learning baselines adapted to the domain of event sequences.
In the semi-supervised setting, where the number of labeled data is limited, our method
demonstrates confident performance: the lesser is the labeled data, the larger is
the performance margin between CoLES and the supervised-only methods.

Finally, we demonstrate superior performance of CoLES in several production-level
applications, internally used in our financial services company.
The proposed method of generating embeddings appears to be useful in production environments
since pre-calculated embeddings can be easily used for different downstream tasks \emph{without}
performing complex and time-consuming computations on the raw event data.


\bibliographystyle{ACM-Reference-Format}
\bibliography{sigmod2022}

\end{document}